\documentclass[10pt]{article}

\usepackage[margin=0.8in]{geometry}
\usepackage{amsmath}
\usepackage{amssymb}
\usepackage{booktabs}
\usepackage{graphicx}
\usepackage{hyperref}
\usepackage{xcolor}

\hypersetup{
  colorlinks=true,
  linkcolor=blue,
  citecolor=blue,
  urlcolor=blue
}
\title{Brain-to-Image Generation: Reconstructing Visual Stimuli from EEG
using Generative Adversarial Networks}
\author{
  Harshit Goyal\\
  BITS Pilani, India\\
  \texttt{2023ac05905@wilp.bits-pilani.ac.in}\\
  \texttt{harshit.goyal.in@gmail.com}
}
\date{}

\begin{document}
\maketitle

\begin{abstract}
Reconstructing visual stimuli from electroencephalography (EEG) is difficult
because scalp measurements have high temporal but limited spatial resolution,
and paired EEG--image datasets remain small relative to modern generative-model
training corpora. We present a reproducible single-subject baseline on
THINGS-EEG2 that first tests the more defensible question of whether EEG can
retrieve the viewed stimulus in a visual embedding space. A compact
temporal--spatial convolutional encoder maps repetition-averaged EEG
($63\times250$) to provided 512-dimensional ViT-B/32 image features. Model
selection uses a concept-disjoint validation split, and final evaluation uses
the official 200-image, 200-concept test gallery. Across three training seeds, the model obtains
$12.83\pm0.58$\%, $39.17\pm1.76$\%, and $58.00\pm1.73$\% image recall at 1,
5, and 10 (mean $\pm$ sample standard deviation), compared with analytical
chance levels of 0.5\%, 2.5\%, and 5.0\%. A session-balanced ablation shows
that averaging more test repetitions generally improves ranking. Applying the
Subject 01 model to the other nine subjects without adaptation causes a sharp
performance drop, exposing subject specificity. We further report exploratory
stress tests of direct conditional generators trained without external visual
weights. Single-subject and ten-subject variants produce noise-dominated
outputs; early decreases in validation $\ell_1$ reverse after one to four
epochs. Finally, we distinguish direct reconstruction from semantic rendering
with a pretrained diffusion prior. The results support reliable coarse
semantic decoding under a closed-set, repetition-averaged protocol, but do not
support faithful recovery of stimulus pixels. Code is available at
\url{https://github.com/harshit-goyal/brain-to-image}.
\end{abstract}

\section{Introduction}

Visual decoding from brain activity asks whether neural measurements recorded
during perception retain enough information to identify or reconstruct the
viewed content. Electroencephalography is attractive for this purpose because
it is non-invasive, portable, inexpensive relative to functional magnetic
resonance imaging, and temporally precise. Its low spatial resolution, volume
conduction, and sensitivity to artifacts nevertheless make image-level
decoding particularly challenging.

Recent work has combined EEG encoders with semantically structured vision
representations and powerful pretrained generators
\cite{song2024nice,li2024visual_decoding,bai2024dreamdiffusion}. This strategy
can produce visually compelling samples, but the generator contributes
substantial prior knowledge. Consequently, a plausible image is not by itself
evidence that stimulus-specific shape, texture, pose, or background was
recovered from EEG.

This work isolates two questions. First, can a compact EEG encoder recover
enough information to rank the correct stimulus above unrelated candidates
under concept-disjoint evaluation? Second, can the available paired data train
a useful image generator from random initialization? The first question yields
a positive retrieval result; the second yields a documented negative result.
This separation prevents image realism supplied by a pretrained prior from
being mistaken for neural reconstruction fidelity.

Our contributions are:
\begin{itemize}
  \item a compact and reproducible EEG-to-visual-feature retrieval pipeline for
  THINGS-EEG2;
  \item concept-disjoint validation, official 200-concept testing, analytical
  chance levels, and a shuffled-pair negative control;
  \item three-seed, session-balanced repetition, and cross-subject robustness
  analyses; and
  \item transparent, compute-bounded failure cases for direct single- and
  multi-subject conditional generation trained without pretrained visual
  weights.
\end{itemize}

\subsection{Research questions and scope}

We organize the study around three research questions:
\begin{enumerate}
  \item[\textbf{RQ1}] Does repetition-averaged EEG retrieve an unseen visual
  concept above closed-gallery chance?
  \item[\textbf{RQ2}] How sensitive is retrieval to the number of EEG
  repetitions, random initialization, and subject shift?
  \item[\textbf{RQ3}] Under the available image and compute budget, what
  failure modes arise when the visual generator is trained from random
  initialization?
\end{enumerate}
RQ1 is the primary confirmatory experiment. RQ2 consists of robustness
analyses, while RQ3 is an exploratory stress test. We do not claim open-world,
single-trial, cross-subject, or faithful pixel reconstruction.

\section{Related Work}

\paragraph{EEG visual decoding.}
Convolutional networks can learn temporal and spatial EEG representations
directly from sensor-time arrays \cite{schirrmeister2017deep}. EEGNet
\cite{lawhern2018eegnet} introduced a compact design based on temporal,
depthwise spatial, and separable convolutions, making it suitable for regimes
with limited neural data. Early visual-decoding studies reported
classification and generation from EEG representations
\cite{spampinato2017humanmind,kavasidis2017brain2image,
palazzo2017gan_brain,palazzo2021multimodal}. Subsequent analysis showed that
block-wise acquisition and trial-level splitting can introduce temporal
leakage and greatly inflate performance \cite{li2019training_test_set}.
Leakage arising from non-independent EEG segments has also been documented
more broadly \cite{brookshire2024leakage}. These findings motivate explicit
definition of the generalization unit and concept-disjoint evaluation.

\paragraph{Large-scale natural-image EEG.}
THINGS-EEG2 provides EEG responses to a broad set of natural object concepts
under rapid visual presentation \cite{gifford2022things_eeg2}. NICE aligns EEG
and image representations contrastively and evaluates zero-shot recognition
on held-out concepts \cite{song2024nice}. Related work uses EEG embeddings and
guided diffusion for retrieval and reconstruction
\cite{li2024visual_decoding}. Our study follows this embedding-space view but
focuses on a compact, inspectable baseline and explicit controls rather than
state-of-the-art comparison.

\paragraph{Visual priors and generation.}
CLIP learns a semantically organized multimodal representation from
large-scale image--text supervision \cite{radford2021clip}. Latent diffusion
models similarly acquire extensive visual knowledge from large image corpora
\cite{rombach2022latentdiffusion}; adversarial diffusion distillation enables
few-step synthesis \cite{sauer2023add}. DreamDiffusion combines masked EEG
pretraining, CLIP alignment, and a pretrained diffusion prior
\cite{bai2024dreamdiffusion}. Such systems are valuable for visualization, but
their output must be interpreted as a combination of decoded neural evidence
and generator prior knowledge.

\section{Data and Protocol}

\subsection{Dataset}

We use THINGS-EEG2 \cite{gifford2022things_eeg2} through the derived,
float16 release distributed with the Uncertainty-Aware Blur Prior (UBP)
implementation \cite{wu2025ubp}. The UBP preprocessing epochs raw data from
$-200$ to 1,000\,ms, applies baseline correction through stimulus onset,
resamples to 250\,Hz, and retains the final 250 samples (0--1,000\,ms).
Multivariate noise normalization is computed separately by session using only
training-partition covariance and is then applied to training and test data.
The primary experiment uses Subject 01. Each EEG observation contains 63
channels and 250 temporal samples. The training partition contains 16,540
unique images from 1,654 concepts and four EEG repetitions per image. The
official test partition contains 200 images from 200 held-out concepts and 80
repetitions per image. The derived release also provides normalized
512-dimensional CLIP ViT-B/32 image features \cite{radford2021clip}.

For the primary experiment, all available repetitions are averaged before
encoding: four during training and 80 during testing. This asymmetry raises
test-time signal-to-noise ratio and means the reported results do not measure
single-trial decoding. No new human participants were recruited, and this work
uses only the released, anonymized dataset under its original data-governance
and ethics procedures.

\begin{table}[t]
  \centering
  \caption{Primary Subject 01 data used in this study. Repetitions are averaged
  into one encoder input per image.}
  \label{tab:data}
  \begin{tabular}{lrrr}
    \toprule
    Partition & Images & Concepts & Repetitions/image\\
    \midrule
    Optimization & 14,890 & 1,489 & 4\\
    Validation & 1,650 & 165 & 4\\
    Official test & 200 & 200 & 80\\
    \bottomrule
  \end{tabular}
\end{table}

\begin{figure}[t]
  \centering
  \includegraphics[width=\linewidth]{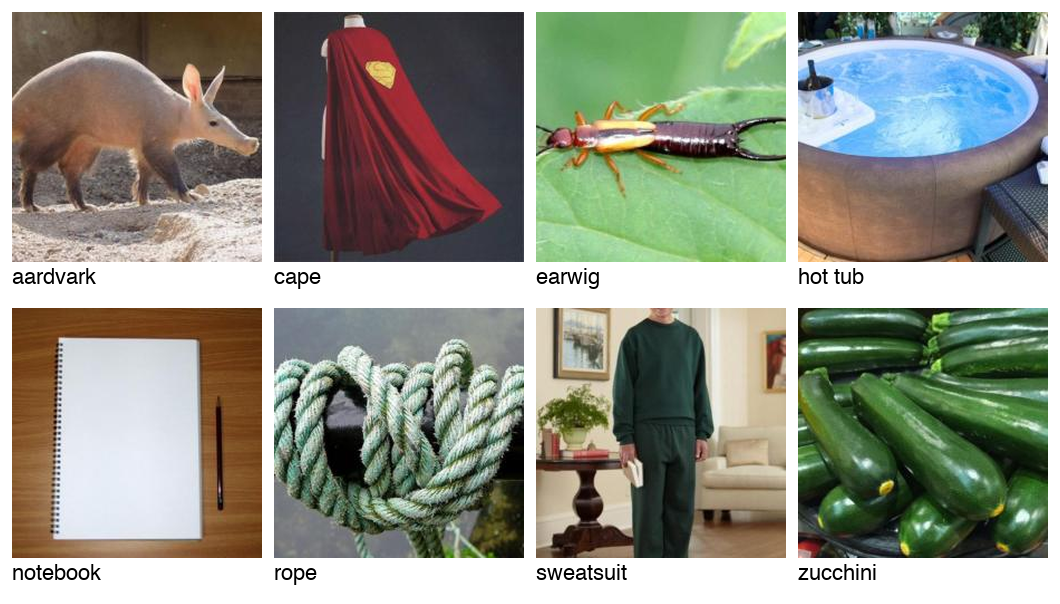}
  \caption{Representative stimulus images from eight distinct concepts in the
  Subject 01 optimization partition. The complete optimization set contains
  14,890 images spanning 1,489 concepts; this montage illustrates its visual
  diversity rather than a cherry-picked performance subset.}
  \label{fig:training-samples}
\end{figure}

\subsection{Leakage-aware split}

The official test set remains untouched during training and model selection.
Within the training partition, concepts rather than trials or image instances
are randomly divided with seed 42: 1,489 concepts (14,890 images) are used for
optimization and 165 concepts (1,650 images) for validation. Thus, no
validation concept appears in training. This controls semantic memorization in
addition to the trial/block leakage concerns identified in prior EEG work
\cite{li2019training_test_set,brookshire2024leakage}.

\begin{figure}[t]
  \centering
  \includegraphics[width=\linewidth]{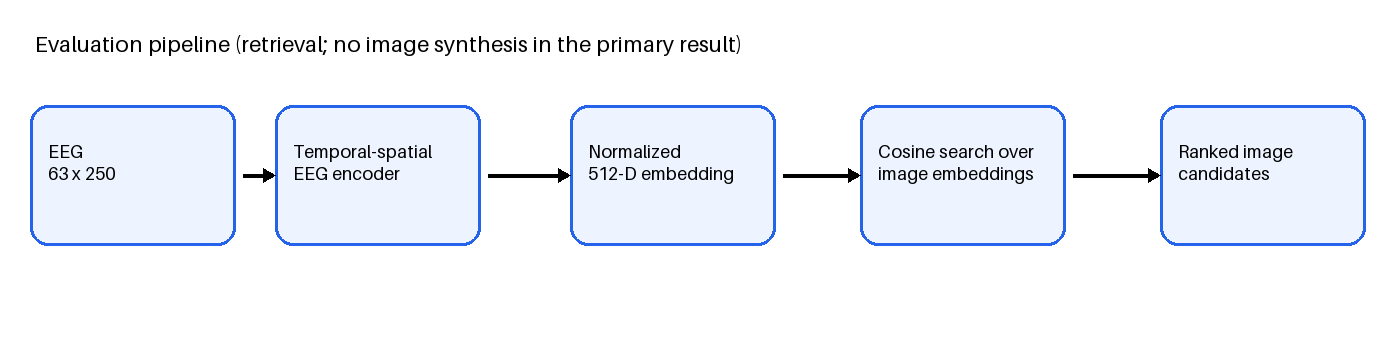}
  \caption{Primary evaluation pipeline. The EEG encoder predicts a normalized
  visual feature, which is compared by cosine similarity with candidate image
  features. Image synthesis is not part of the primary retrieval result.}
  \label{fig:pipeline}
\end{figure}

\section{Method}

\subsection{EEG encoder}

The encoder follows the compact temporal--spatial pattern of EEGNet
\cite{lawhern2018eegnet}. Input $x\in\mathbb{R}^{63\times250}$ first passes
through a temporal convolution with 32 filters and kernel width 25. A grouped
spatial convolution spans all 63 channels and produces 64 feature maps. After
batch normalization, GELU activation, average pooling, and dropout, a
depthwise temporal convolution of width 15 and a pointwise convolution produce
128 maps. A second pooling stage leaves 15 temporal positions. The flattened
representation is projected to 512 dimensions, layer-normalized, and
$\ell_2$-normalized.

\begin{table}[t]
  \centering
  \caption{EEG encoder architecture. The model has 999,008 trainable
  parameters; 98.6\% are in the final projection.}
  \label{tab:architecture}
  \begin{tabular}{llll}
    \toprule
    Stage & Operation & Output channels & Temporal width\\
    \midrule
    Input & EEG tensor & 63 & 250\\
    Temporal & Conv2D, kernel $1\times25$ & 32 & 250\\
    Spatial & grouped Conv2D, kernel $63\times1$ & 64 & 250\\
    Pool 1 & average pool $1\times4$ & 64 & 62\\
    Separable & depthwise $1\times15$, pointwise & 128 & 62\\
    Pool 2 & average pool $1\times4$ & 128 & 15\\
    Projection & flatten, linear, layer norm & 512 & 1\\
    \bottomrule
  \end{tabular}
\end{table}

\subsection{Alignment objective}

Let $\hat{z}_i$ denote the normalized EEG prediction and $z_i$ the normalized
visual feature for item $i$. For a batch of size $B$, similarities are
\[
  s_{ij} = \frac{\hat{z}_i^\top z_j}{\tau},
\]
where $\tau=0.07$. We optimize symmetric cross-entropy over EEG-to-image and
image-to-EEG directions, plus cosine alignment:
\[
  \mathcal{L} =
  \frac{1}{2}\left[
    \operatorname{CE}(S,I)+\operatorname{CE}(S^\top,I)
  \right]
  + 0.25\left(1-\frac{1}{B}\sum_i \hat{z}_i^\top z_i\right).
\]
Images sharing a concept are masked as off-diagonal negatives so the objective
does not explicitly push semantically equivalent images apart.

The model is trained for 12 epochs with AdamW, learning rate
$3\times10^{-4}$, weight decay $10^{-3}$, batch size 128, dropout 0.35, and
gradient-norm clipping at 1.0. Temporal kernels span 100\,ms and 60\,ms at the
250\,Hz sampling rate. The fixed 12-epoch budget and remaining hyperparameters
were chosen before test evaluation. The checkpoint with the highest validation
concept recall at 5 is selected. The final epoch was selected in each of the
three seeds and validation had not fully saturated, so the runs should be
treated as baselines rather than optimized estimates.

\begin{table}[t]
  \centering
  \caption{Primary retrieval training configuration.}
  \label{tab:hyperparameters}
  \begin{tabular}{lr@{\qquad}lr}
    \toprule
    Setting & Value & Setting & Value\\
    \midrule
    Optimizer & AdamW & Batch size & 128\\
    Learning rate & $3\times10^{-4}$ & Weight decay & $10^{-3}$\\
    Epoch budget & 12 & Dropout & 0.35\\
    Temperature & 0.07 & Cosine weight & 0.25\\
    Gradient clip & 1.0 & Validation fraction & 0.10\\
    \bottomrule
  \end{tabular}
\end{table}

\subsection{Retrieval evaluation}

At test time, each EEG prediction is compared with all 200 image features by
cosine similarity. Recall at $k$ is one when the paired test image appears in
the top $k$ candidates. Because the official gallery has one image per
concept, image and concept recall are identical on this split. Analytical
chance is $k/200$. We additionally apply one deterministic random permutation
to EEG--stimulus correspondences and report the resulting shuffled control.
Because only one permutation is used, this control is expected to fluctuate
around analytical chance and is not a permutation-test estimate.

\paragraph{Metric definitions.}
For query $i$, let $r_i$ be the one-indexed rank of its paired image among the
gallery. Recall at $k$ is
\[
  R@k = \frac{1}{N}\sum_{i=1}^{N}\mathbb{1}[r_i\leq k].
\]
We additionally report mean and median rank and paired cosine similarity.
Wilson intervals treat the 200 test concepts as Bernoulli observations. The
shared candidate gallery introduces dependence, so intervals are descriptive
rather than a complete hierarchical uncertainty model.

\section{Results}

\begin{table}[t]
  \centering
  \caption{Subject 01 retrieval on the official 200-concept test gallery.
  Confidence intervals are 95\% Wilson intervals over 200 test concepts.}
  \label{tab:retrieval}
  \begin{tabular}{lrrrr}
    \toprule
    Metric & Model & 95\% CI & Chance & Shuffled EEG\\
    \midrule
    Recall@1  & 13.5\% & [9.4, 18.9]  & 0.5\% & 0.5\%\\
    Recall@5  & 41.0\% & [34.4, 47.9] & 2.5\% & 2.0\%\\
    Recall@10 & 59.0\% & [52.1, 65.6] & 5.0\% & 3.5\%\\
    \bottomrule
  \end{tabular}
\end{table}

\begin{figure}[t]
  \centering
  \includegraphics[width=0.96\linewidth]{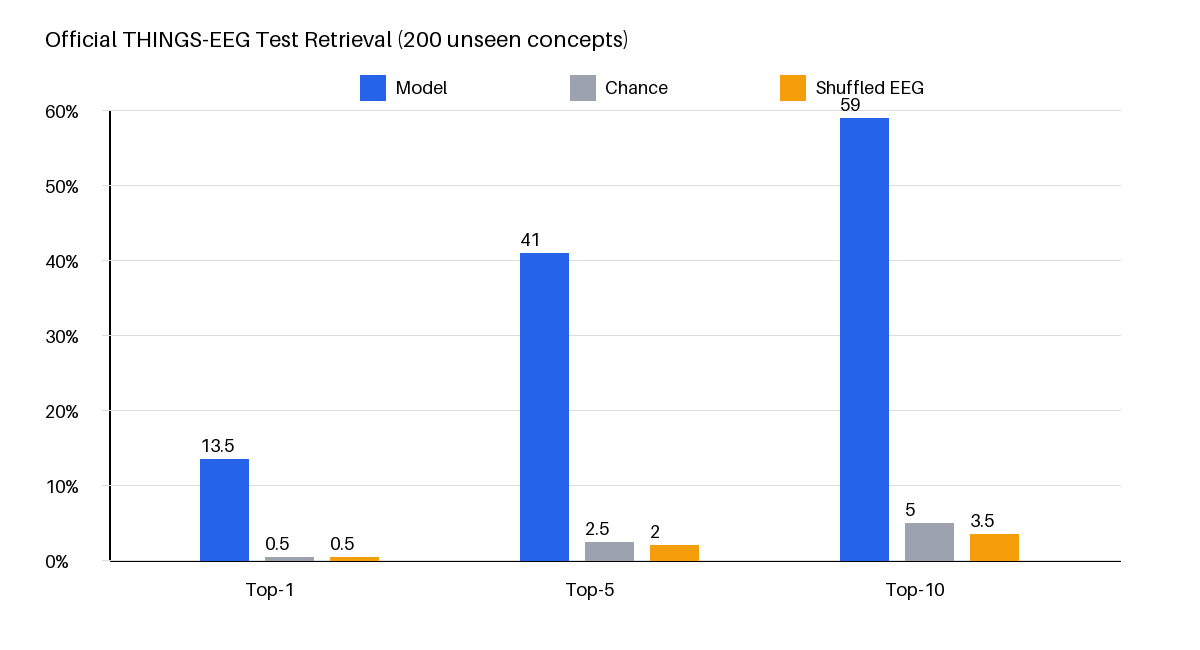}
  \caption{Retrieval accuracy substantially exceeds analytical chance and a
  fixed shuffled-EEG control at each value of $k$.}
  \label{fig:retrieval}
\end{figure}

Table~\ref{tab:retrieval} and Figure~\ref{fig:retrieval} summarize the primary
result. The correct image is ranked first for 27 of 200 concepts, within the
top five for 82, and within the top ten for 118. The median rank is 8 and the
mean rank is 17.11. Mean paired cosine similarity is 0.219.

Under an independent-query binomial approximation, recall at 1 is above chance
with $p<10^{-20}$. The Wilson intervals and this tail probability treat queries
as independent, an approximation because all queries share the same gallery.
These calculations establish performance above random gallery selection under
the stated test protocol; they do not establish
cross-subject generalization or rule out every dataset confound.

\subsection{Training behavior}

Concept recall at 5 on the validation partition rises from 4.97\% after the
first epoch to 21.39\% after epoch 12, compared with analytical chance of
approximately 3.00\%. Validation image recall at 5 reaches 8.12\%, compared
with image-level chance of approximately 0.30\%. The
divergence between image and concept retrieval on validation is expected
because each validation concept has multiple image instances.

\subsection{Robustness analyses}

To address RQ2, we repeat the complete Subject 01 training protocol with seeds
7 and 21 alongside the original seed 42. Table~\ref{tab:seeds} shows limited
run-to-run variation. Recall@1 ranges from 12.5\% to 13.5\%, and all runs have
a median rank of 8 or 9. The aggregate standard deviations are small relative
to the gap from analytical chance, although three seeds remain insufficient
for a full characterization of optimization uncertainty.

\begin{table}[t]
  \centering
  \caption{Official Subject 01 test results across independently initialized
  training runs. The last row reports mean $\pm$ sample standard deviation.}
  \label{tab:seeds}
  \begin{tabular}{lrrrr}
    \toprule
    Seed & R@1 & R@5 & R@10 & Median rank\\
    \midrule
    42 & 13.5\% & 41.0\% & 59.0\% & 8\\
    7  & 12.5\% & 39.0\% & 59.0\% & 8\\
    21 & 12.5\% & 37.5\% & 56.0\% & 9\\
    \midrule
    Aggregate & $12.83\pm0.58$\% & $39.17\pm1.76$\% &
    $58.00\pm1.73$\% & $8.33\pm0.58$\\
    \bottomrule
  \end{tabular}
\end{table}

\begin{figure}[t]
  \centering
  \includegraphics[width=0.94\linewidth]{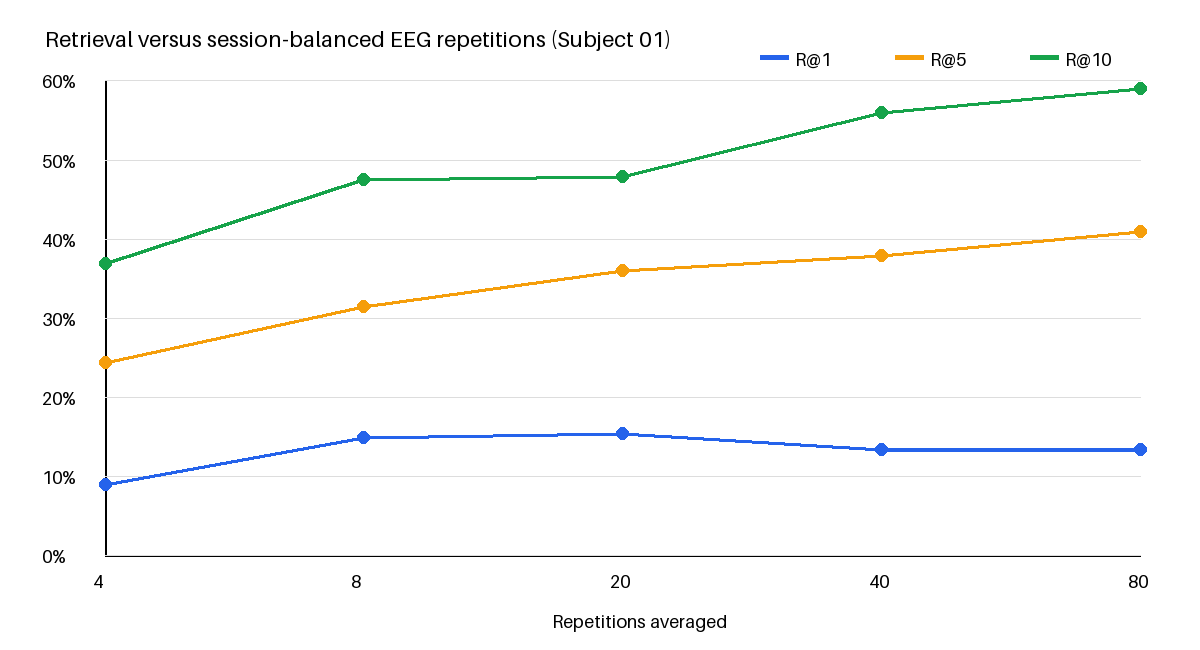}
  \caption{Retrieval versus the number of session-balanced test repetitions.
  Higher-$k$ recall improves overall as averaging suppresses trial noise.}
  \label{fig:repetitions}
\end{figure}

We next hold the seed-42 weights fixed and vary test-time averaging. The 80
test repetitions comprise four sessions of 20 repetitions. To avoid
confounding repetition count with session coverage, each ablation averages an
equal number of the earliest repetitions from every session. Consequently,
the tested totals are 4, 8, 20, 40, and 80 rather than a prefix of the stored
trial axis. As Table~\ref{tab:repetitions} and
Figure~\ref{fig:repetitions} show, mean and median rank improve consistently
with additional repetitions. Recall@5 and recall@10 also rise overall, from
24.5\% and 37.0\% with four repetitions to 41.0\% and 59.0\% with 80.
Recall@1 is non-monotonic, peaking at 15.5\% with 20 repetitions; this
illustrates that top-1 alone is a noisy summary for a 200-query test set.

\begin{table}[t]
  \centering
  \caption{Session-balanced repetition ablation for the fixed seed-42 model.
  Each condition contributes equally from all four test sessions.}
  \label{tab:repetitions}
  \begin{tabular}{rrrrrr}
    \toprule
    Repetitions & R@1 & R@5 & R@10 & Median rank & Mean rank\\
    \midrule
    4  & 9.0\%  & 24.5\% & 37.0\% & 19 & 35.87\\
    8  & 15.0\% & 31.5\% & 47.5\% & 12 & 26.28\\
    20 & 15.5\% & 36.0\% & 48.0\% & 11 & 20.45\\
    40 & 13.5\% & 38.0\% & 56.0\% & 9  & 17.76\\
    80 & 13.5\% & 41.0\% & 59.0\% & 8  & 17.11\\
    \bottomrule
  \end{tabular}
\end{table}

\begin{figure}[t]
  \centering
  \includegraphics[width=0.94\linewidth]{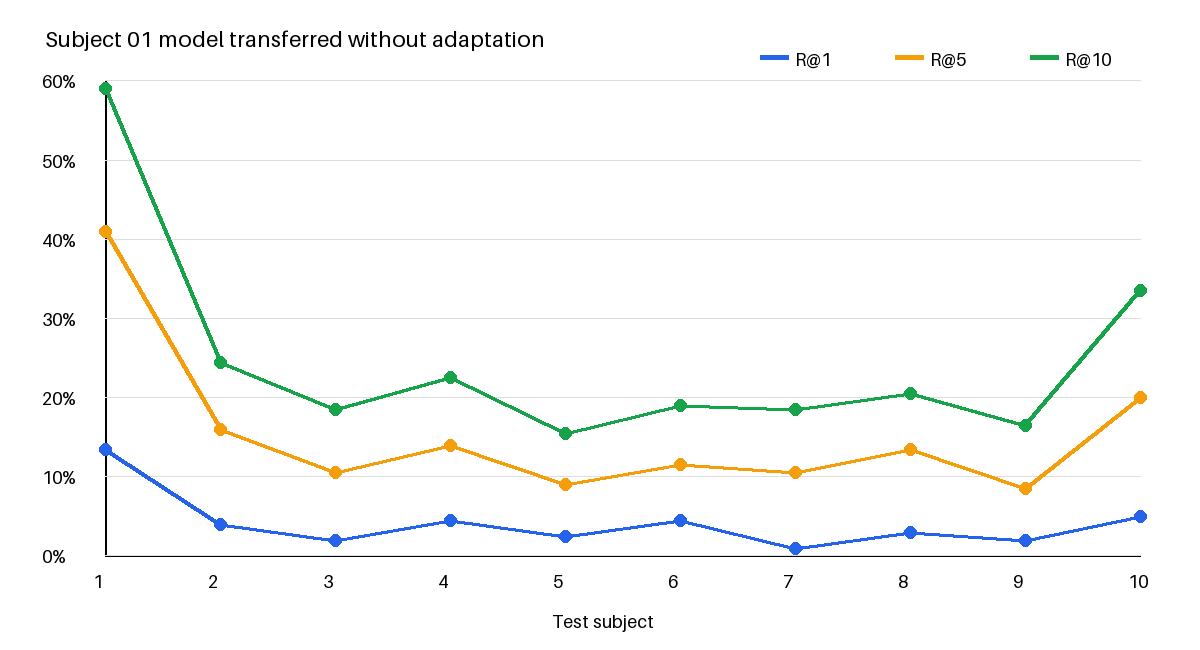}
  \caption{Zero-adaptation transfer of the Subject 01 encoder. Performance
  drops sharply for every other participant, indicating subject-specific
  neural-to-visual mappings.}
  \label{fig:crosssubject}
\end{figure}

Finally, we apply the seed-42 Subject 01 model without fine-tuning to the
official test data of Subjects 02--10, averaging all 80 repetitions. This
strict transfer test retains the shared channel layout and preprocessing but
does not estimate a subject-adapted mapping. Table~\ref{tab:crosssubject} and
Figure~\ref{fig:crosssubject} show a large transfer penalty: recall@10 falls
from 59.0\% on Subject 01 to 15.5--33.5\% on the other subjects, while median
rank worsens from 8 to 21--48. Every transferred subject remains numerically
above analytical chance at recall@10, but these exploratory values should not
be read as inferential evidence without subject-level training replications
and uncertainty estimates.

\begin{table}[t]
  \centering
  \caption{Subject 01 model evaluated without adaptation on all subjects.
  Analytical chance is 0.5\%, 2.5\%, and 5.0\% at ranks 1, 5, and 10.}
  \label{tab:crosssubject}
  \begin{tabular}{rrrrr@{\qquad}rrrrr}
    \toprule
    Subj. & R@1 & R@5 & R@10 & Med. & Subj. & R@1 & R@5 & R@10 & Med.\\
    \midrule
    1 & 13.5 & 41.0 & 59.0 & 8  & 6  & 4.5 & 11.5 & 19.0 & 39\\
    2 & 4.0  & 16.0 & 24.5 & 28 & 7  & 1.0 & 10.5 & 18.5 & 42\\
    3 & 2.0  & 10.5 & 18.5 & 48 & 8  & 3.0 & 13.5 & 20.5 & 40\\
    4 & 4.5  & 14.0 & 22.5 & 31 & 9  & 2.0 & 8.5  & 16.5 & 46\\
    5 & 2.5  & 9.0  & 15.5 & 40 & 10 & 5.0 & 20.0 & 33.5 & 21\\
    \bottomrule
  \end{tabular}
\end{table}

\section{Generation Stress Tests}

We evaluated generation separately from retrieval to avoid conflating visual
plausibility with decoded information.

\begin{figure}[t]
  \centering
  \includegraphics[width=\linewidth]{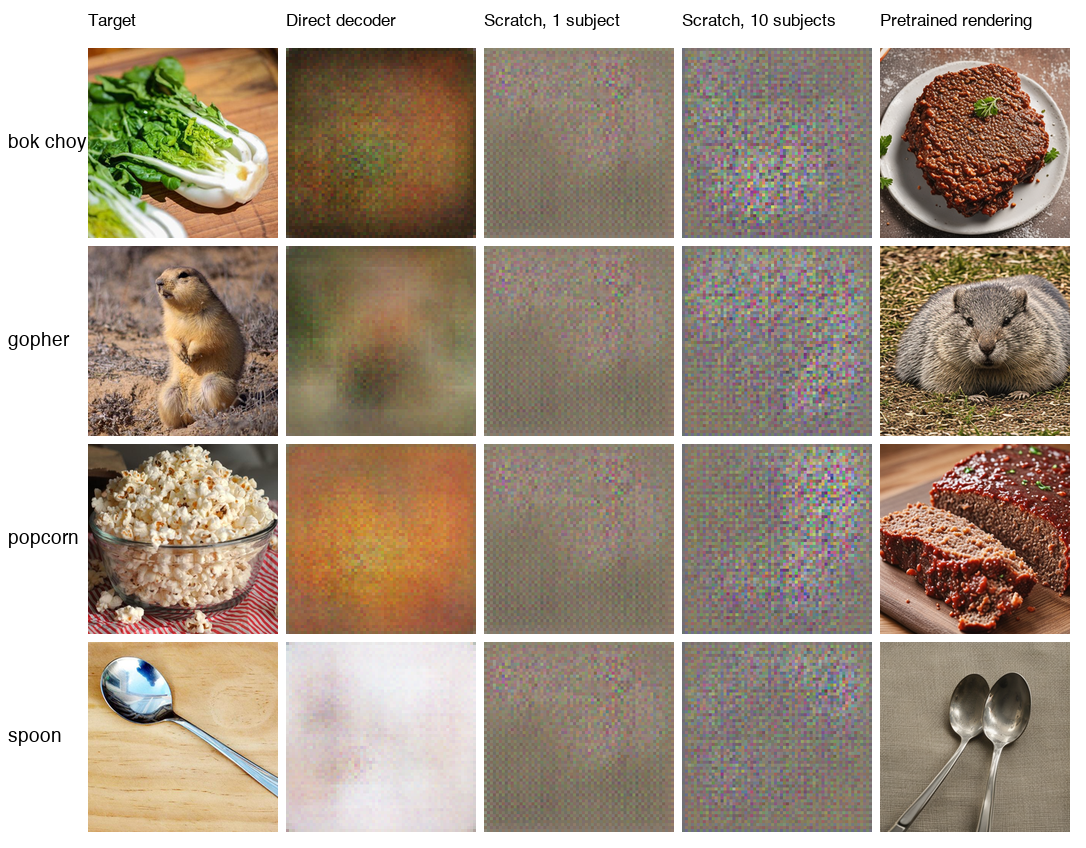}
  \caption{Qualitative outputs for four held-out test stimuli. The direct
  feature-to-pixel decoder preserves only coarse color and layout. Both
  generators trained from random initialization are noise-dominated. The
  pretrained rendering is sharp because a large diffusion prior receives the
  EEG-retrieved class label; it is a semantic visualization, not recovery of
  the target image.}
  \label{fig:generation-comparison}
\end{figure}

\paragraph{Feature-to-pixel decoder.}
A transposed-convolution decoder trained to map provided visual features to
$64\times64$ pixels reached a best concept-disjoint validation
multi-scale reconstruction loss of 0.1715. When driven by predicted EEG
features, outputs preserved broad color and low-frequency layout but were
blurred and lacked recognizable object identity. This exposes a distribution
mismatch between true visual features used to train the decoder and noisy EEG
predictions used at inference.

\paragraph{Single-subject conditional GAN.}
We jointly trained a temporal--spatial EEG encoder, transposed-convolution
generator, and projection discriminator from random initialization, using no
external visual weights. The six-epoch run used Adam, batch size 256,
generator and discriminator learning rates of $2\times10^{-4}$ and
$10^{-4}$, and an $\ell_1$ weight of 10. Validation $\ell_1$ error improved
to 0.4442 at epoch 4 and then degraded. A training-set mean-image baseline
achieves 0.4382 on the same validation partition, so the generator does not
beat this trivial baseline. Across a fixed eight-image test sample, mean
pairwise output RMSE is 0.00048, versus 0.3395 among the corresponding target
images (all scaled to $[0,1]$). Generated outputs are therefore nearly
invariant and noise-dominated under this training budget.

\paragraph{Ten-subject conditional GAN.}
To test whether additional neural observations resolve collapse, we trained a
subject-conditioned variant across all ten locally available subjects. Subject
files were streamed sequentially, with 1,500 images sampled per subject and
epoch from the same at-most 14,890-image training pool. Four epochs exposed
the model to 60,000 subject--image examples while retaining four averaged
repetitions per image.
Validation $\ell_1$ was best after epoch 1 (0.4480) and subsequently degraded
to 0.4565. Mean pairwise output RMSE is 0.0822 on the same fixed eight-image
sample, still far below target diversity of 0.3395. Outputs remained
noise-dominated. This compute-bounded stress test suggests that adding subjects
did not compensate for limited unique visual examples or adversarial
instability under our setup. It does not imply that all direct EEG-conditioned
generators must fail.

\paragraph{Pretrained diffusion visualization.}
For qualitative demonstration only, predicted top-1 concepts were converted
to text prompts and rendered with SD-Turbo, a distilled pretrained diffusion
model \cite{sauer2023add}. These images are sharp because the diffusion model
supplies a large external visual prior. They are semantic renderings of the
decoded class, not direct EEG-to-pixel reconstructions, and are excluded from
the quantitative claims. The only EEG-derived signal passed to the generator
is one discrete class selected from 200 candidates (at most
$\log_2 200\approx7.6$ bits); no continuous EEG or image feature enters the
diffusion model.

\section{Discussion}

The retrieval result shows that repetition-averaged Subject 01 EEG contains
information that generalizes to unseen THINGS-EEG2 concepts in a closed
200-candidate gallery. The gap between recall and chance is large, and the
shuffled control is consistent with random correspondence. The experiment
therefore supports coarse semantic decoding under this protocol.

The robustness analyses qualify that result. Similar performance across three
initializations suggests that the finding is not driven by one fortunate
seed. Session-balanced averaging improves rank and higher-$k$ recall, showing
that much of the reported accuracy depends on repeated measurements rather
than single-trial signal quality. Cross-subject transfer is consistently much
weaker than within-subject performance. A deployable decoder would therefore
need subject-specific calibration or an explicitly trained subject-invariant
representation; neither is demonstrated here.

The generation experiments produce a complementary conclusion. Pixel losses
encourage average-looking images, while adversarial training with limited
paired data is unstable. Increasing the number of subjects adds neural
observations but not new image content: the ten-subject experiment draws from
the same at-most 14,890-image training pool and only ten images per concept.
Learning both a broad natural-image distribution and an EEG-conditioned
mapping from this corpus proved unsuccessful.

A stronger future design would train a discrete image autoencoder or
generative prior on a substantially larger image corpus, freeze the visual
decoder, and then align EEG with its latent representation. If the visual
model is pretrained, evaluation should separately quantify what information
comes from EEG, for example through retrieval, ablations, shuffled controls,
and comparisons against class-only prompts.

\section{Limitations}

The primary training result is subject-specific and uses only Subject 01.
Zero-adaptation transfer to other subjects degrades sharply and is not a
substitute for independently trained within-subject models. Eighty test
repetitions are averaged for each stimulus, and the repetition ablation starts
at four trials to preserve equal session representation; performance should
not be interpreted as real-time or single-trial decoding. Evaluation is closed-set:
the model ranks a known gallery rather than identifying an arbitrary image
from the open world. The study uses provided ViT-B/32 features and therefore
inherits their pretrained visual semantics. Three training seeds reduce but
do not eliminate optimization uncertainty, and only one shuffled permutation
is reported. The repetition conditions reuse the same test concepts and use
the earliest trials per session rather than resampling all possible subsets.
Hyperparameters were selected for a practical baseline rather than through a
comprehensive search. Finally, direct image generation did not succeed;
visual realism from the auxiliary diffusion experiment comes primarily from
the pretrained generator.

\section{Ethical Considerations}

Brain-decoding results are vulnerable to sensational interpretation. The
present system cannot recover private thoughts, operate without a controlled
visual-stimulus protocol, or reconstruct arbitrary images. Its strongest
result uses 80 repeated test presentations and a known 200-image gallery.
Reporting the work as ``mind reading'' would therefore be inaccurate.
Potential future applications should require informed consent, clear limits
on secondary use, secure handling of neural recordings, and evaluation across
demographic and physiological variation. Because this study performs
secondary analysis of an anonymized public dataset, it introduces no new
participant intervention.

\section{Conclusion}

A compact EEG encoder can retrieve unseen visual concepts from the official
THINGS-EEG2 test gallery substantially above chance when repeated trials are
averaged, with similar results across three training seeds. Session-balanced
analysis confirms a strong dependence on repeated measurements, while
zero-adaptation evaluation reveals substantial cross-subject degradation. In
contrast, direct generators trained from scratch on the available paired
images collapse or produce blurred, noise-dominated outputs. The key
methodological conclusion is that semantic EEG decoding and photorealistic
image synthesis must be evaluated separately. Retrieval provides evidence
about information present in EEG, whereas a powerful image prior determines
much of the appearance of generated samples.

\section*{Reproducibility}

Training, evaluation, and analysis code given the derived UBP-preprocessed
release, along with commands, dependency specifications, tests, configurations,
and result artifacts, are available at
\url{https://github.com/harshit-goyal/brain-to-image}. Large dataset files and
model checkpoints are excluded from version control. The repository records
the random seed, architecture, split procedure, analytical baselines, and
fixed shuffled control used for the reported experiment.

\appendix
\section{Implementation and Reproduction Details}

Experiments were executed with Python 3.9, PyTorch 2.8, and NumPy 2.0 on a
MacBook Air with an Apple M2 processor (8 CPU cores), 16\,GB unified memory,
and the PyTorch MPS backend. Dataset files are loaded as float16 and converted
to float32 per sample before repetition averaging. Data-loader workers remain
at zero on macOS to avoid copying the approximately 2\,GB in-memory subject
array into spawned workers.

The main commands are:
\begin{verbatim}
python3 main.py train --subject 1 --epochs 12 \
  --batch-size 128 --seed 42 --device mps
python3 main.py evaluate --checkpoint outputs/sub-01/best.pt \
  --split test --device mps
python3 paper/analyze_robustness.py --device mps
python3 paper/analyze_seeds.py --device mps
\end{verbatim}

Each reported output directory includes a JSON configuration, epoch history,
and evaluation metrics. Dataset tensors, downloaded model weights, and
checkpoints are omitted from Git because of size; paths and expected tensor
shapes are documented in the repository README.


\begin{thebibliography}{10}

\bibitem{bai2024dreamdiffusion}
Yunpeng Bai, Xintao Wang, Yan-pei Cao, Yixiao Ge, Chun Yuan, and Ying Shan.
\newblock {DreamDiffusion}: High-quality {EEG}-to-image generation with
  temporal masked signal modeling and {CLIP} alignment.
\newblock In {\em Computer Vision -- ECCV 2024}, pages 472--488. Springer,
  2024.

\bibitem{brookshire2024leakage}
Geoffrey Brookshire, Jake Kasper, Nicholas~M. Blauch, Yunan~Charles Wu, Ryan
  Glatt, David~A. Merrill, Spencer Gerrol, Keith~J. Yoder, Colin Quirk, and
  Ch{\'e} Lucero.
\newblock Data leakage in deep learning studies of translational {EEG}.
\newblock {\em Frontiers in Neuroscience}, 18:1373515, 2024.

\bibitem{gifford2022things_eeg2}
Alessandro~T. Gifford, Kshitij Dwivedi, Gemma Roig, and Radoslaw~M. Cichy.
\newblock A large and rich {EEG} dataset for modeling human visual object
  recognition.
\newblock {\em NeuroImage}, 264:119754, 2022.

\bibitem{kavasidis2017brain2image}
Isaak Kavasidis, Simone Palazzo, Concetto Spampinato, Daniela Giordano, and
  Mubarak Shah.
\newblock {Brain2Image}: Converting brain signals into images.
\newblock In {\em Proceedings of the 25th ACM International Conference on
  Multimedia}, pages 1809--1817. ACM, 2017.

\bibitem{lawhern2018eegnet}
Vernon~J. Lawhern, Amelia~J. Solon, Nicholas~R. Waytowich, Stephen~M. Gordon,
  Chou~P. Hung, and Brent~J. Lance.
\newblock {EEGNet}: A compact convolutional neural network for {EEG}-based
  brain--computer interfaces.
\newblock {\em Journal of Neural Engineering}, 15(5):056013, 2018.

\bibitem{li2024visual_decoding}
Dongyang Li, Chen Wei, Shiying Li, Jiachen Zou, and Quanying Liu.
\newblock Visual decoding and reconstruction via {EEG} embeddings with guided
  diffusion.
\newblock In {\em Advances in Neural Information Processing Systems},
  volume~37, pages 102822--102864, 2024.

\bibitem{li2019training_test_set}
Ren Li, Jared~S. Johansen, Hamad Ahmed, Thomas~V. Ilyevsky, Ronnie~B. Wilbur,
  Hari~M. Bharadwaj, and Jeffrey~Mark Siskind.
\newblock Training on the test set? an analysis of spampinato et al.
\newblock {\em arXiv preprint arXiv:1812.07697}, 2019.

\bibitem{palazzo2021multimodal}
Simone Palazzo, Concetto Spampinato, Isaak Kavasidis, Daniela Giordano, Joseph
  Schmidt, and Mubarak Shah.
\newblock Decoding brain representations by multimodal learning of neural
  activity and visual features.
\newblock {\em IEEE Transactions on Pattern Analysis and Machine Intelligence},
  43(11):3833--3849, 2021.

\bibitem{palazzo2017gan_brain}
Simone Palazzo, Concetto Spampinato, Isaak Kavasidis, Daniela Giordano, and
  Mubarak Shah.
\newblock Generative adversarial networks conditioned by brain signals.
\newblock In {\em 2017 IEEE International Conference on Computer Vision}, pages
  3430--3438. IEEE, 2017.

\bibitem{radford2021clip}
Alec Radford, Jong~Wook Kim, Chris Hallacy, Aditya Ramesh, Gabriel Goh,
  Sandhini Agarwal, Girish Sastry, Amanda Askell, Pamela Mishkin, Jack Clark,
  Gretchen Krueger, and Ilya Sutskever.
\newblock Learning transferable visual models from natural language
  supervision.
\newblock In {\em Proceedings of the 38th International Conference on Machine
  Learning}, volume 139 of {\em Proceedings of Machine Learning Research},
  pages 8748--8763. PMLR, 2021.

\bibitem{rombach2022latentdiffusion}
Robin Rombach, Andreas Blattmann, Dominik Lorenz, Patrick Esser, and Bj{\"o}rn
  Ommer.
\newblock High-resolution image synthesis with latent diffusion models.
\newblock In {\em 2022 IEEE/CVF Conference on Computer Vision and Pattern
  Recognition}, pages 10684--10695. IEEE, 2022.

\bibitem{sauer2023add}
Axel Sauer, Dominik Lorenz, Andreas Blattmann, and Robin Rombach.
\newblock Adversarial diffusion distillation.
\newblock {\em arXiv preprint arXiv:2311.17042}, 2023.

\bibitem{schirrmeister2017deep}
Robin~Tibor Schirrmeister, Jost~Tobias Springenberg, Lukas Dominique~Josef
  Fiederer, Martin Glasstetter, Katharina Eggensperger, Michael Tangermann,
  Frank Hutter, Wolfram Burgard, and Tonio Ball.
\newblock Deep learning with convolutional neural networks for {EEG} decoding
  and visualization.
\newblock {\em Human Brain Mapping}, 38(11):5391--5420, 2017.

\bibitem{song2024nice}
Yonghao Song, Bingchuan Liu, Xiang Li, Nanlin Shi, Yijun Wang, and Xiaorong
  Gao.
\newblock Decoding natural images from {EEG} for object recognition.
\newblock In {\em International Conference on Learning Representations}, 2024.

\bibitem{spampinato2017humanmind}
Concetto Spampinato, Simone Palazzo, Isaak Kavasidis, Daniela Giordano, Nasim
  Souly, and Mubarak Shah.
\newblock Deep learning human mind for automated visual classification.
\newblock In {\em 2017 IEEE Conference on Computer Vision and Pattern
  Recognition}, pages 4503--4511. IEEE, 2017.

\bibitem{wu2025ubp}
Haitao Wu, Qing Li, Changqing Zhang, Zhen He, and Xiaomin Ying.
\newblock Bridging the vision-brain gap with an uncertainty-aware blur prior.
\newblock In {\em Proceedings of the IEEE/CVF Conference on Computer Vision and
  Pattern Recognition}, pages 2246--2257, 2025.

\end{thebibliography}
\end{document}